\documentclass[11pt]{article}

\usepackage[utf8]{inputenc}
\usepackage[T1]{fontenc}
\usepackage[margin=1in]{geometry}
\usepackage{amsmath,amssymb,amsfonts}
\usepackage{graphicx}
\graphicspath{{figs/}{./}}
\usepackage{booktabs}
\usepackage{float}
\usepackage{multirow}
\usepackage{siunitx}
\usepackage{xcolor}
\usepackage{subcaption}
\usepackage{enumitem}
\usepackage[numbers,sort&compress]{natbib}
\usepackage[hidelinks]{hyperref}
\usepackage{cleveref}

\usepackage{tikz}
\usetikzlibrary{positioning,arrows.meta,shapes.geometric,fit,backgrounds,calc}

\newcommand{\sys}{SFGA}
\newcommand{\Sys}{SFGA}
\newcommand{\dset}{\mathcal{D}}          
\newcommand{\axis}{a}                    
\newcommand{\est}{\hat{s}}               
\newcommand{\ci}{\mathrm{CI}}            
\newcommand{\ciw}[1]{w_{\ci,#1}}         
\newcommand{\thr}{\tau}                  
\newcommand{\gold}{g^{\star}}            
\newcommand{\dec}{\delta}                
\newcommand{\rec}{\rho}                  
\newcommand{\prob}{\hat{p}}             
\DeclareMathOperator{\Div}{div}
\DeclareMathOperator{\Util}{util}
\DeclareMathOperator{\Red}{red}

\definecolor{inputgreen}{RGB}{198,232,197}
\definecolor{measorange}{RGB}{250,219,178}
\definecolor{statsblue}{RGB}{197,220,246}
\definecolor{gatered}{RGB}{247,199,199}
\definecolor{judgepurple}{RGB}{221,209,242}
\definecolor{verdictpurple}{RGB}{206,190,236}
\definecolor{edgegray}{RGB}{90,90,90}

\title{SFGA: A Statistics-First Gating Architecture with Adjudicative\\
       Escalation for Trustworthy SFT Data Procurement}
\author{
  Arther Tian\textsuperscript{a},
  Alex Ding\textsuperscript{a,*},
  Frank Chen\textsuperscript{a}\\
  Simon Wu\textsuperscript{a},
  Aaron Chan\textsuperscript{a}\\
  \textsuperscript{a}DGrid AI\\[0.5em]
  \textsuperscript{*}Corresponding author: \texttt{alex.ding@dgrid.ai}
}
\date{}

\begin{document}
\maketitle

\begin{abstract}
Procuring supervised fine-tuning (SFT) data forces a buyer to decide, before
any downstream training, whether a candidate corpus is worth acquiring. We
present \sys{}, a statistics-first gating architecture that treats procurement as a
cost-aware routing problem over three intrinsic quality axes---diversity,
utility, and redundancy. Cheap blind measurements are summarised into
per-axis estimates with confidence intervals; a gate accepts a decision only
when intervals are tight, sample sizes are adequate, and the axes agree,
otherwise it escalates the case to an adjudicative debate between a
buy-advocate and a reject-advocate judge, resolved by a presiding verdict.
On a controlled benchmark of 12 datasets ($2{\times}3{\times}2$
grid over the three axes) with 5 seeds, the gate reaches
0.90 accuracy and 0.83 $F_1$ at \$0.017 per unit,
sitting between an always-verify baseline (0.75) and an oracle upper bound
(0.98) while spending less than always-escalate (\$0.020). We further report
honest negative diagnostics of the debate path: a con-side win rate of
0.80 ($p\approx3{\times}10^{-6}$) and a 52\% position-flip
rate under advocate swapping expose negativity and positional biases that a
naive LLM-judge would hide. We frame the injected-knob evaluation explicitly
as a controlled synthetic benchmark for measurement fidelity and routing
calibration, and delimit external validity as future work.
\end{abstract}

\section{Introduction}
\label{sec:intro}


Supervised fine-tuning (SFT)~\citep{ouyang2022instructgpt,wei2022flan}
increasingly runs on data a team did not create
itself: corpora purchased from vendors, scraped and repackaged by third
parties, or pooled across an organization. Whoever pays for such data must
decide, before committing to a fine-tuning run, whether a candidate corpus is
worth acquiring at all~\citep{ghorbani2019datashapley,jia2019datavaluation}---a
judgment made under a budget, on a corpus that is
usually too large to read and too opaque to trust. The decision is
two-sided: acquire a corpus that quietly degrades the model and the buyer
loses both money and a training cycle; pass over a genuinely useful corpus
and the opportunity is gone. Procurement, not post-hoc filtering, is where
the money is committed, and it is the layer we target.

Three families of methods bear on this judgment, each with a characteristic
failure. Human audit is the most trusted---careful human curation of even a
small corpus can suffice for alignment~\citep{zhou2023lima}---but it scales
poorly and cannot be afforded at corpus volume. A single-shot LLM
judge~\citep{zheng2023mtbench,liu2023geval} scales but is costly per
unit at scale and, as we quantify in \Cref{sec:results}, is biased and poorly
calibrated: its verdict swings with the order in which alternatives are
presented~\citep{wang2024notfair} and skews systematically toward rejection. Statistics over cheap
intrinsic measurements is the least expensive option and, on clear-cut cases,
is genuinely decisive; but at the boundaries---when an estimate straddles a
threshold, when the sample is too small to support a conclusion, or when
quality axes disagree---it commits confidently to answers it has no basis for.

We take these methods to be complementary rather than competing, and argue
that the scarce resource---expensive adjudication---should be spent only where
cheap statistics are demonstrably insufficient. \Sys{} measures three
intrinsic quality axes of a candidate corpus---diversity, utility, and
redundancy---summarizes each as an interval-valued estimate, and consults a
gate. The gate commits to a statistics-only verdict only when the intervals
are tight, the samples adequate, and the axes unanimous; every ambiguous,
borderline, or conflicting case is escalated to an adjudicative
debate~\citep{du2024debate,irving2018debate} between
a buy-advocate and a reject-advocate, resolved by a presiding verdict. The
same escalation path that rescues hard cases is also where judge pathologies
live, so we audit it directly---via advocate swapping---rather than trusting
its verdicts at face value. While this cheap-first, escalate-when-uncertain,
cost-aware philosophy is shared with a broader line of the authors' work on
trustworthy evaluation~\citep{ourpriorwork}, both the problem addressed
here---data procurement---and the mechanism---a statistical gate over
corpus-level quality axes---are new.

\begin{figure}[t]
  \centering
  \resizebox{0.88\linewidth}{!}{%
  \begin{tikzpicture}[font=\small]
    \draw[-{Stealth[length=2mm]},thick] (0,0) -- (9.2,0)
      node[right,font=\footnotesize]{Cost per unit};
    \draw[-{Stealth[length=2mm]},thick] (0,0) -- (0,6.4);
    \node[rotate=90,anchor=south,font=\footnotesize] at (-0.45,3.2) {Reliability};
    \draw[dashed,edgegray] (1.0,1.2) .. controls (4.2,1.7) and (6.0,4.2) .. (7.6,5.6);
    \node[edgegray,font=\scriptsize,align=center] at (5.9,2.5)
      {existing cost--reliability\\ frontier};
    \node[circle,draw,fill=statsblue,minimum size=5mm] (stat) at (1.5,1.5) {};
    \node[right=1mm of stat,align=left,font=\scriptsize]
      {Pure statistics\\ cheap, brittle at boundaries};
    \node[circle,draw,fill=judgepurple,minimum size=5mm] (judge) at (7.1,3.4) {};
    \node[left=1mm of judge,align=right,font=\scriptsize]
      {Single LLM judge\\ costly, biased};
    \node[circle,draw,fill=measorange,minimum size=5mm] (human) at (7.7,5.3) {};
    \node[above=1mm of human,align=center,font=\scriptsize]
      {Human audit\\ trusted, unscalable};
    \node[star,star points=5,draw,fill=gatered,minimum size=8mm,
          inner sep=0pt] (ours) at (3.0,5.0) {};
    \node[above=1mm of ours,align=center,font=\scriptsize]
      {\textbf{\sys{} (ours)}\\ escalate only the hard cases};
  \end{tikzpicture}}
  \caption{Positioning. Existing procurement practice trades cost against
  reliability; \sys{} spends cheaply by default and pays for adjudication
  only on contested cases.}
  \label{fig:positioning}
\end{figure}
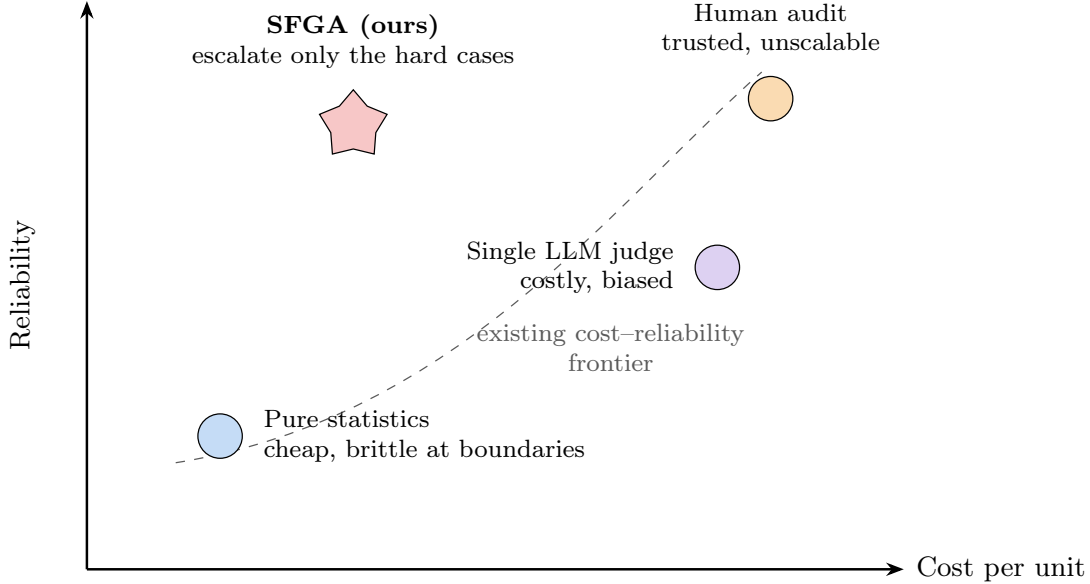

On a controlled benchmark of 12 datasets spanning a
$2{\times}3{\times}2$ grid of injected quality knobs, run across 5
seeds, \sys{} reaches 0.90 routing accuracy at \$0.017 per
unit---close to an oracle that routes with ground truth (0.98) and, notably,
cheaper than escalating every case (\$0.020) while far more accurate than
never escalating (0.75). It is also the best-calibrated strategy we test.
Against a random router matched to its exact budget (0.82), the remaining gap
isolates the value of the routing \emph{decision} from the value of the money
spent. \Cref{fig:positioning} situates \sys{} on the cost--reliability
frontier of existing practice. We frame the injected-knob evaluation
explicitly as a controlled study of measurement fidelity and routing
calibration, and treat external validity as future work (\Cref{sec:discussion}).

\Sys{} makes three contributions:
\begin{itemize}[leftmargin=1.4em,itemsep=2pt]
  \item A statistics-first procurement gate. A decision rule over
  interval-valued estimates of diversity, utility, and redundancy that
  commits to a cheap verdict only when the statistics are decisive and
  unanimous, and otherwise routes to adjudication (\Cref{sec:design}).
  \item Adjudicative escalation with honest diagnostics. A
  buy/reject advocate debate with a presiding verdict, audited for
  positional and negativity bias via advocate swapping---turning judge
  pathologies into measured quantities rather than hidden failure modes
  (\Cref{sec:results}).
  \item A controlled benchmark and cost-matched evaluation. A
  $2{\times}3{\times}2$ synthetic grid with a knob-derived gold label used
  strictly for measurement-fidelity and routing-calibration analysis, plus
  a cost-matched random baseline that isolates the gate's contribution from
  its budget (\Cref{sec:setup,sec:results}).
\end{itemize}

\section{Related Work}
\label{sec:related}

\paragraph{Data quality and procurement.}
Two recent surveys map the fast-growing space of data selection for language
models~\citep{albalak2024surveydataselection,qin2024datatsunami}. A separate
line quantifies the \emph{worth} of data through its marginal contribution to
a trained model, most influentially via the Shapley
value~\citep{ghorbani2019datashapley,jia2019datavaluation}. These valuation
methods answer a different question from ours and at a different price: they
require training or retraining to estimate a datum's contribution, whereas we
decide whether to acquire a corpus \emph{before} any training, from cheap
intrinsic measurements plus selective adjudication.

\paragraph{SFT data selection and filtering.}
A large body of work shows that small, carefully chosen instruction sets can
match or beat much larger ones~\citep{zhou2023lima,chen2024alpagasus}, and
proposes automatic criteria for choosing them: complexity, quality, and
diversity in DEITA~\citep{liu2024deita}, tag-based diversity in
InsTag~\citep{lu2024instag}, instruction-following difficulty in
IFD and Superfiltering~\citep{li2023ifd,li2024superfiltering}, and
learnability-style scores in Instruction
Mining~\citep{cao2023instructionmining}. Redundancy and diversity have their
own tools---exact and semantic deduplication~\citep{lee2022dedup,abbas2023semdedup}
and sentence embeddings for coverage~\citep{reimers2019sbert}---and the whole
enterprise rests on instruction-tuning foundations~\citep{ouyang2022instructgpt,wei2022flan,wang2023selfinstruct}.
The crucial distinction is the unit of decision: these methods score or filter
\emph{individual samples} to assemble a training subset, while we operate at
the \emph{corpus} layer and emit a procurement verdict, governed by a
statistical stopping rule and an escalation path rather than a per-sample
score.

\paragraph{LLM-as-judge and multi-agent debate.}
Using strong models to evaluate outputs is now
standard~\citep{zheng2023mtbench,liu2023geval}, but such judges carry
well-documented biases---their verdicts depend on option order and other
superficial factors~\citep{wang2024notfair}, which directly motivates our
position-swap diagnostic. Debate has been proposed both to improve model
reasoning~\citep{du2024debate} and as an alignment mechanism in which
adversarial advocates surface information for a
judge~\citep{irving2018debate}. We borrow the adversarial structure but put it
to a narrower use---debate is not our default but the escalation path for
contested procurement cases only---and we treat the judge's biases as
quantities to be measured rather than assumed away.

\paragraph{Statistical testing and sequential decision.}
The gate is built from classical tools: the Wilson interval for
proportions~\citep{wilson1927interval}, sequential hypothesis
testing~\citep{wald1945sequential}, the bootstrap for interval
estimation~\citep{efron1979bootstrap}, and the Brier score~\citep{brier1950verification}
and modern calibration analysis~\citep{guo2017calibration} for scoring
probabilistic verdicts. Our contribution is not a new estimator but their
composition into a decision rule that recognizes when statistics are
sufficient and, only otherwise, pays for adjudication.

\section{System Design}
\label{sec:design}

\subsection{Problem setup and notation}
A candidate data package $\dset$ is inspected by scanning a fixed prefix.
For each quality axis $\axis\in\{\Div,\Util,\Red\}$ a blind measurement
yields a point estimate $\est_\axis$ with a confidence interval
$\ci_\axis=[\ell_\axis,h_\axis]$ (width $\ciw{\axis}=h_\axis-\ell_\axis$) over
$n$ observations. A gate emits a path decision
$\dec\in\{\textsf{verify},\textsf{escalate}\}$; the chosen path returns a
procurement recommendation $\rec\in\{\textsf{buy},\textsf{caution},
\textsf{reject},\textsf{ask\_more}\}$ and a calibrated probability $\prob$.
The controlled gold label $\gold$ (\Cref{sec:setup}) is used for offline
evaluation only and never enters the online decision.

\subsection{Main architecture}
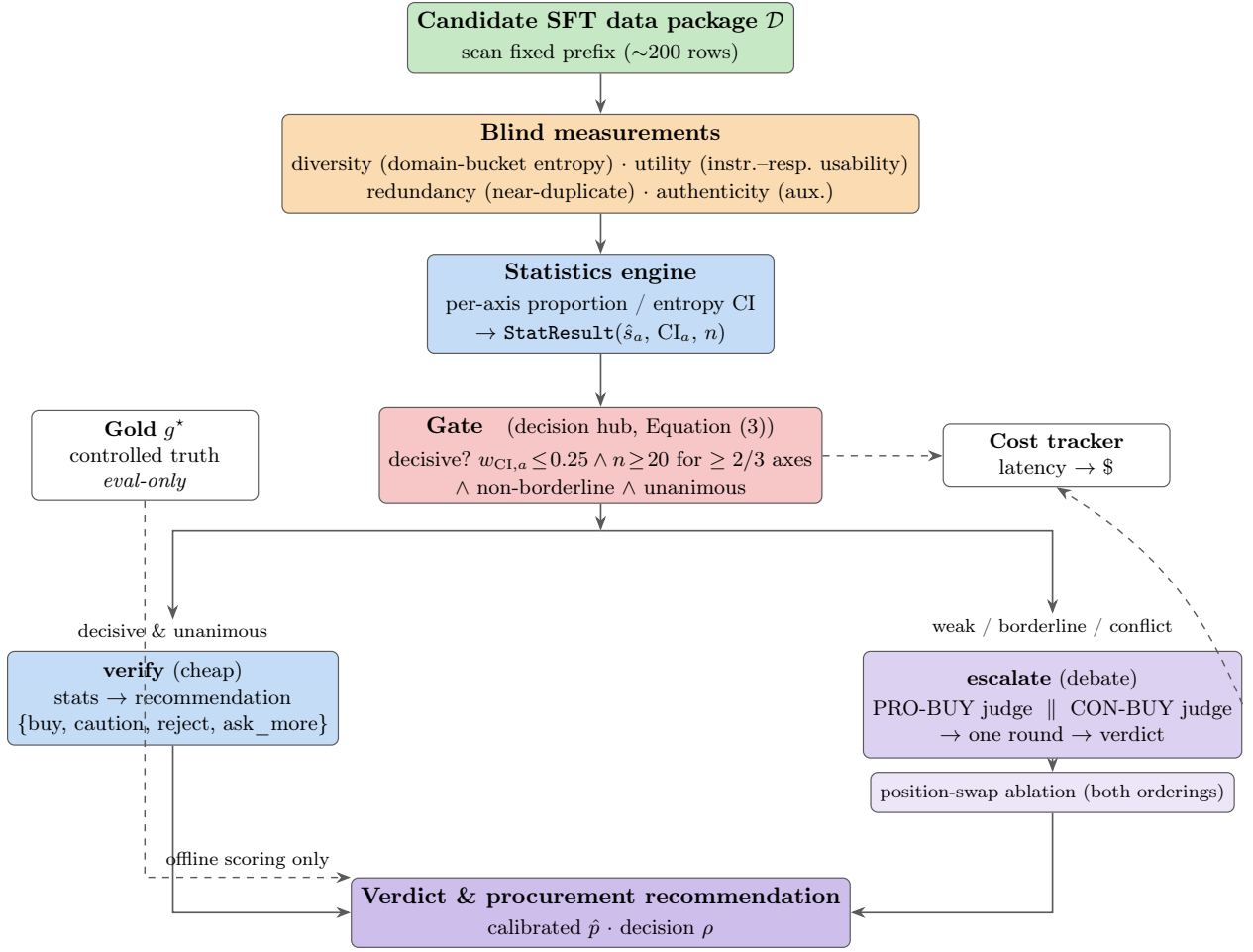
\begin{figure*}[t]
\centering
\resizebox{\textwidth}{!}{%
\begin{tikzpicture}[
  font=\small,
  >={Stealth[length=2.2mm]},
  node distance=6mm and 10mm,
  block/.style={rounded corners=3pt,draw=edgegray,line width=0.5pt,
                align=center,inner sep=4pt,minimum height=8mm},
  stage/.style={block,minimum width=52mm},
  smallb/.style={block,minimum width=34mm,font=\footnotesize},
  flow/.style={->,draw=edgegray,line width=0.7pt},
  feedback/.style={->,draw=edgegray,line width=0.6pt,dashed},
]

\node[stage,fill=inputgreen] (input)
  {\textbf{Candidate SFT data package} $\dset$\\[1pt]
   \footnotesize scan fixed prefix ($\sim$200 rows)};

\node[stage,fill=measorange,below=of input] (meas)
  {\textbf{Blind measurements}\\[1pt]
   \footnotesize diversity (domain-bucket entropy) $\cdot$
   utility (instr.--resp.\ usability)\\
   \footnotesize redundancy (near-duplicate) $\cdot$ authenticity (aux.)};

\node[stage,fill=statsblue,below=of meas] (stats)
  {\textbf{Statistics engine}\\[1pt]
   \footnotesize per-axis proportion / entropy CI\\
   \footnotesize $\rightarrow$ \texttt{StatResult}$(\est_\axis,\,\ci_\axis,\,n)$};

\node[block,fill=gatered,below=8mm of stats,minimum width=66mm] (gate)
  {\textbf{Gate}\quad\footnotesize (decision hub, \Cref{eq:gate})\\[2pt]
   \footnotesize decisive?\ $\ciw{\axis}\!\le\!0.25 \wedge n\!\ge\!20$ for $\ge 2/3$ axes\\
   \footnotesize $\wedge$ non-borderline $\wedge$ unanimous};

\node[smallb,fill=statsblue,below left=22mm and 6mm of gate] (verify)
  {\textbf{verify} \footnotesize(cheap)\\[1pt]
   \footnotesize stats $\rightarrow$ recommendation\\
   \footnotesize \{buy, caution, reject, ask\_more\}};

\node[smallb,fill=judgepurple,below right=22mm and 6mm of gate,minimum height=16mm] (escalate)
  {\textbf{escalate} \footnotesize(debate)\\[2pt]
   \footnotesize PRO-BUY judge $\;\|\;$ CON-BUY judge\\
   \footnotesize $\rightarrow$ one round $\rightarrow$ verdict};

\node[smallb,fill=judgepurple!55,below=2mm of escalate,minimum height=6mm,font=\scriptsize] (swap)
  {position-swap ablation (both orderings)};

\node[stage,fill=verdictpurple,below=56mm of gate,minimum width=74mm] (verdict)
  {\textbf{Verdict \& procurement recommendation}\\[1pt]
   \footnotesize calibrated $\prob$ $\cdot$ decision $\rec$};

\node[smallb,fill=white,right=18mm of gate,font=\footnotesize] (cost)
  {\textbf{Cost tracker}\\ \footnotesize latency $\rightarrow$ \$};
\node[smallb,fill=white,left=18mm of gate,font=\footnotesize] (goldbox)
  {\textbf{Gold} $\gold$\\ \footnotesize controlled truth\\ \footnotesize\emph{eval-only}};

\node[font=\scriptsize,fill=white,inner sep=1.5pt,above=1.5mm of verify] (lblv)
  {decisive \& unanimous};
\node[font=\scriptsize,fill=white,inner sep=1.5pt,above=1.5mm of escalate] (lble)
  {weak / borderline / conflict};

\draw[flow] (input) -- (meas);
\draw[flow] (meas)  -- (stats);
\draw[flow] (stats) -- (gate);
\draw[flow] (gate.south) -- ++(0,-4mm) coordinate (branch);
\draw[flow] (branch) -| (lblv);
\draw[flow] (branch) -| (lble);
\draw[flow] (verify.south)   |- (verdict.west);
\draw[flow] (escalate.south) -- (swap.north);
\draw[flow] (swap.south)     |- (verdict.east);

\draw[feedback] (gate.east) -- (cost.west);
\draw[feedback] (escalate.east) to[bend right=18] (cost.south);
\draw[feedback] (goldbox.south) |- (verdict.north west)
  node[pos=0.75,above,font=\scriptsize]{offline scoring only};

\end{tikzpicture}}
\caption{End-to-end architecture. Cheap blind measurements feed a statistics
engine whose interval-valued estimates drive a gate (red, the
decision hub). Decisive and unanimous cases take the cheap \textsf{verify}
path; weak, borderline, or conflicting cases escalate to an
\textsf{adjudicative debate} between buy- and reject-advocates with a
presiding verdict. The controlled gold label $\gold$ (left) is used for
offline scoring only and never enters the online decision.}
\label{fig:arch}
\end{figure*}

\Cref{fig:arch} shows the end-to-end pipeline.

\subsection{The gate rule}
An axis estimate is \emph{decisive} when it has adequate sample size and a
tight interval:
\begin{equation}
  \text{decisive}(\axis) \;=\; \big[\, n_\axis \ge n_{\min}\,\big]\;\wedge\;
  \big[\, \ciw{\axis} \le w_{\max}\,\big].
  \label{eq:decisive}
\end{equation}
A decisive axis is classified relative to its threshold $\thr_\axis$ with a
borderline band of half-width $m$:
\begin{equation}
  \mathrm{side}(\axis)=
  \begin{cases}
    \textsf{borderline} & \ell_\axis<\thr_\axis<h_\axis \text{ or } |\est_\axis-\thr_\axis|\le m,\\
    \textsf{pass}       & \text{clears } \thr_\axis \text{ in the good direction},\\
    \textsf{fail}       & \text{otherwise.}
  \end{cases}
  \label{eq:side}
\end{equation}
The gate takes the cheap \textsf{verify} path iff a fraction $f$ of axes are
decisive, none are borderline, and the decisive axes are unanimous;
otherwise it escalates:
\begin{equation}
  \dec=
  \begin{cases}
    \textsf{verify}   & |\{\axis:\text{decisive}\}|\ge f\,|A|,\ \text{none borderline, all agree},\\[2pt]
    \textsf{escalate} & \text{weak CI, borderline, or pass/fail conflict.}
  \end{cases}
  \label{eq:gate}
\end{equation}
We use $n_{\min}{=}20$, $w_{\max}{=}0.25$, $f{=}2/3$, $m{=}0.05$, thresholds
$\thr_{\Div}{=}0.15$, $\thr_{\Util}{=}0.55$, $\thr_{\Red}{=}0.25$, and
$z{=}1.96$ intervals (all in \Cref{tab:hparams}).

\subsection{Adjudicative escalation}
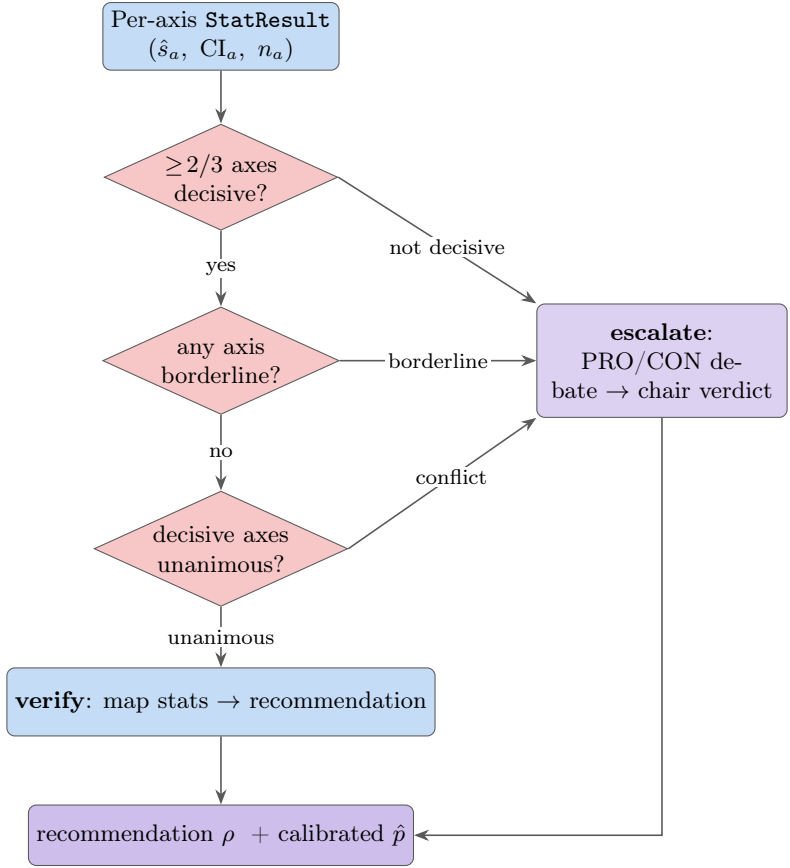
\begin{figure}[t]
  \centering
  \begin{tikzpicture}[
    font=\footnotesize,
    >={Stealth[length=2mm]},
    io/.style={rounded corners=3pt,draw=edgegray,align=center,
               inner sep=3pt,minimum height=8mm},
    dec/.style={diamond,draw=edgegray,fill=gatered,aspect=2.2,
                align=center,inner sep=1pt,minimum height=1.3cm},
    proc/.style={draw=edgegray,align=center,inner sep=3pt,minimum height=9mm},
    fl/.style={->,draw=edgegray,line width=0.6pt},
    lbl/.style={font=\scriptsize,fill=white,inner sep=1pt},
  ]
    \node[io,fill=statsblue] (start) at (0,0)
      {Per-axis \texttt{StatResult}\\ $(\est_\axis,\ \ci_\axis,\ n_\axis)$};
    \node[dec,below=7mm of start] (d1)
      {$\ge\!2/3$ axes\\ decisive?};
    \node[dec,below=10mm of d1] (d2) {any axis\\ borderline?};
    \node[dec,below=10mm of d2] (d3) {decisive axes\\ unanimous?};
    \node[proc,rounded corners=3pt,fill=statsblue,below=8mm of d3] (verify)
      {\textbf{verify}: map stats $\rightarrow$ recommendation};
    \node[io,fill=verdictpurple,below=9mm of verify] (out)
      {recommendation $\rec$ \ +\ calibrated $\prob$};
    \node[proc,rounded corners=3pt,fill=judgepurple,right=26mm of d2,
          minimum height=1.5cm,text width=3.1cm] (esc)
      {\textbf{escalate}: PRO/CON debate $\rightarrow$ chair verdict};

    \draw[fl] (start) -- (d1);
    \draw[fl] (d1) -- (d2) node[lbl,midway]{yes};
    \draw[fl] (d2) -- (d3) node[lbl,midway]{no};
    \draw[fl] (d3) -- (verify) node[lbl,midway]{unanimous};
    \draw[fl] (verify) -- (out);

    \draw[fl] (d1.east) -- (esc.north west) node[lbl,pos=0.55]{not decisive};
    \draw[fl] (d2.east) -- (esc.west) node[lbl,pos=0.5]{borderline};
    \draw[fl] (d3.east) -- (esc.south west) node[lbl,pos=0.55]{conflict};
    \draw[fl] (esc.south) |- (out.east);
  \end{tikzpicture}
  \caption{Gate decision flow. The single most load-bearing figure: the gate
  is a branch, not a waterfall.}
  \label{fig:gateflow}
\end{figure}

When the gate returns \textsf{escalate}---because intervals are wide, an
estimate is borderline, or the axes disagree---the case is handed to an
adjudicative debate rather than to a single judge. Two advocates receive the
\emph{same} statistical evidence, including which metrics are flagged as weak,
but are assigned opposing and fixed stances: a buy-advocate argues that the
corpus is acceptable to purchase, and a reject-advocate argues against
acquisition and stresses risk. Each advocate first states an opening argument
confined to intrinsic quality---diversity and coverage, instruction--response
utility, and within-pack redundancy---and then, in a single rebuttal round,
answers the opponent's opening. A presiding chair reads both openings and both
rebuttals alongside the statistics and returns a procurement recommendation
$\rec\in\{\textsf{buy},\textsf{caution},\textsf{reject},\textsf{ask\_more}\}$
together with a calibrated buy-probability $\prob$. \Cref{fig:gateflow} traces
the full branch.

The escalation path is at once the most expensive component and the one most
exposed to the documented biases of LLM judges, so we instrument it rather
than trust it. For every escalated case we additionally run the debate with
the two advocate roles swapped in presentation order; a verdict that changes
under this swap reflects dependence on position rather than on evidence. We
report the resulting positional and negativity diagnostics in
\Cref{sec:results} and treat them as first-class results rather than caveats:
the purpose of a statistics-first design is precisely to keep this fallible
path off the critical route for the cases that cheap statistics already
settle.

\section{Experimental Setup}
\label{sec:setup}

\paragraph{Controlled benchmark.}
We construct a $2{\times}3{\times}2$ grid of injected quality knobs on top of
Dolly-15k~\citep{conover2023dolly}: a diversity mode (low/high), a utility
degradation rate ($0/25/50\%$), and a redundancy duplication rate
($0/30\%$), giving 12 datasets of a few hundred instruction--response pairs
each. Every knob has a known ground-truth value. Diversity is set by sampling
across nine domain buckets, so the high mode approaches uniform coverage; the
utility rate is the fraction of pairs deliberately corrupted, so a $25\%$ rate
fixes $\Util_{\text{true}}=0.75$; and the redundancy rate is the fraction of
near-duplicates injected. Each dataset is run over 5 seeds, yielding 60 routing
units.

\paragraph{Gold label (offline only).}
The procurement gold is a deterministic function of the injected knobs: a
corpus should be bought iff its true diversity, utility, and (low) redundancy
all clear fixed thresholds,
$\Div_{\text{true}}\ge0.15 \wedge \Util_{\text{true}}\ge0.55 \wedge
\Red_{\text{true}}\le0.25$. This label defines correctness for accuracy,
$F_1$, and calibration, but it never enters the online decision---the gate and
both paths see only the blind measurements. Because the same knobs that define
the gold are also what the measurements recover, we treat the benchmark as a
controlled test of \emph{measurement fidelity and routing calibration} rather
than of downstream external validity, a distinction we return to in
\Cref{sec:discussion}.

\paragraph{Blind measurements.}
Each axis is estimated from a scanned prefix with no access to the knobs:
diversity as the entropy of the empirical distribution over the nine domain
buckets, utility as the pass rate of an instruction--response usability check,
and redundancy as the near-duplicate rate; an auxiliary authenticity rate is
also computed. Every estimate carries an interval and a sample count, as
formalized in \Cref{sec:design}.

\paragraph{Judges and cost.}
Escalated cases invoke three LLM roles---the buy- and reject-advocates and the
presiding chair (\Cref{sec:design})---drawn from a single model family. All
calls are real, and the per-unit cost is the token-derived dollar cost
accumulated along a unit's path. Restricting to one model family and one base
corpus is a deliberate scope choice whose limits we state in
\Cref{sec:discussion}.

\paragraph{Metrics and baselines.}
We score each strategy by routing accuracy and $F_1$ against the gold label,
by mean per-unit cost in dollars, and by the calibration (expected calibration
error and Brier score) of its self-reported buy-probability.
\Cref{tab:baselines} lists the strategies we compare, from a cheap
statistics-only floor to an oracle upper bound, including a random router held
to the gate's exact budget so that the routing decision can be separated from
the money spent.
\begin{table}[t]
  \centering
  \caption{Baselines and evaluators (roles). Full config in
  \Cref{tab:hparams}.}
  \label{tab:baselines}
  \begin{tabular}{@{}lll@{}}
    \toprule
    Method & Mechanism & Role \\
    \midrule
    \sys{}              & stats-first route $\rightarrow$ verify/escalate & proposed \\
    always\_verify      & statistics only, never escalate                 & cheap floor \\
    always\_escalate    & debate on every unit                            & expensive ceiling \\
    oracle\_route       & routes with gold                                & upper bound \\
    random\_cost\_matched & random path at matched budget                 & budget control \\
    stats\_only         & statistics with no recommendation mapping       & ablation \\
    direct\_llm         & single-shot judge, no debate                    & weak baseline \\
    \bottomrule
  \end{tabular}
\end{table}

\section{Results}
\label{sec:results}

\subsection{Measurement fidelity (controlled)}
\begin{figure}[t]
  \centering
  \includegraphics[width=\linewidth]{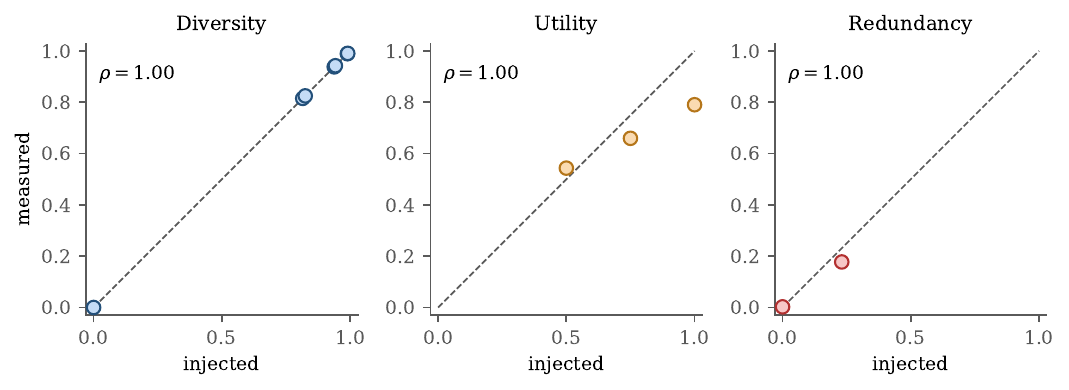}
  \caption{Measurement fidelity on the controlled bench: blind estimates vs.\
  injected ground truth for each axis, with the identity line. Recovery is
  near-perfect by construction and is a fidelity check, not external validity.}
  \label{fig:fidelity}
\end{figure}

Everything downstream depends on the blind measurements being faithful
proxies for the injected knobs, so we check this first. Across all three axes
the measured estimates recover the injected values almost perfectly (Spearman
$\rho\approx1.00$; \Cref{fig:fidelity}). We read this conservatively: because
the measurements and
the gold label are both functions of the same injected knobs, near-perfect
recovery is close to an identity by construction. It establishes that the
measurement layer is not the bottleneck on this bench---a necessary
sanity check---but it is not, and we do not present it as, evidence of
downstream external validity (\Cref{sec:discussion}).

\subsection{Routing}
\begin{figure}[t]
  \centering
  \includegraphics[width=0.72\linewidth]{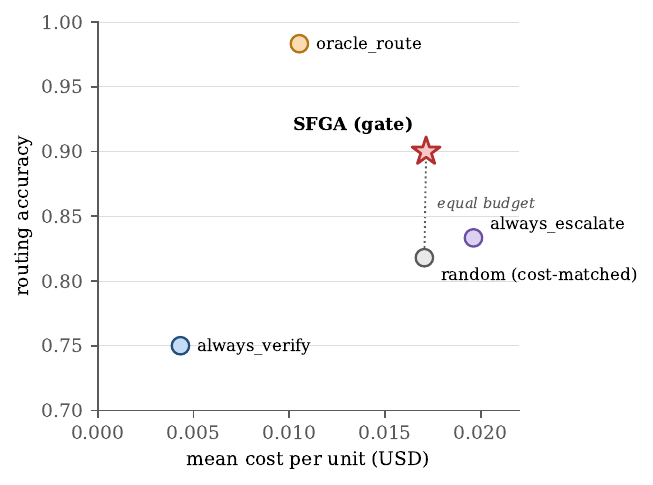}
  \caption{Routing accuracy vs.\ mean per-unit cost. The gate (star) sits near
  the oracle in accuracy while spending less than always-escalate, and beats a
  random router held to its exact budget (dotted, equal-budget guide).}
  \label{fig:routing}
\end{figure}
\Cref{tab:routing} and \Cref{fig:routing} are the headline comparison. Routing the blind measurements
through the gate reaches 0.90 accuracy and 0.83 $F_1$ at \$0.017 per unit. Two
comparisons carry the result. First, against the cheap always-verify floor
(0.75), escalation buys a fifteen-point accuracy gain: the cases the gate
forwards are exactly the ones statistics alone gets wrong. Second, and less
obviously, escalating \emph{every} case is worse on both axes---always-escalate
scores 0.83 accuracy and only 0.69 $F_1$ at a higher \$0.020---because
indiscriminate debate drags the biased adjudication path (quantified in
\Cref{subsec:debate}) onto cases the statistics had already settled correctly.
The gate thus captures most of the oracle's accuracy (0.98) without ever seeing
the gold, and its margin over a random router held to its exact budget (0.82 at
the same \$0.017) shows the gain comes from \emph{where} it spends, not merely
how much.
\begin{table}[t]
  \centering
  \caption{Routing results. The gate approaches the oracle at a fraction of
  the always-escalate detour, and beats the cost-matched random baseline.}
  \label{tab:routing}
  \sisetup{table-format=1.3}
  \begin{tabular}{@{}l S S S[table-format=1.4] c@{}}
    \toprule
    Strategy & {Acc.} & {$F_1$} & {Cost (\$)} & $n$ \\
    \midrule
    \sys{} (gate)        & \bfseries 0.900 & \bfseries 0.833 & 0.0171 & 60 \\
    always\_verify       & 0.750 & 0.727 & 0.0043 & 60 \\
    always\_escalate     & 0.833 & 0.687 & 0.0196 & 60 \\
    oracle\_route        & 0.983 & 0.976 & 0.0105 & 60 \\
    random\_cost\_matched & 0.818 & {--} & 0.0171 & 60 \\
    stats\_only          & 0.750 & 0.727 & 0.0000 & 36 \\
    direct\_llm          & 0.500 & 0.308 & 0.0000 & 36 \\
    \bottomrule
  \end{tabular}
\end{table}

On the routing units the gate escalates 50 of the 60 and keeps 10 on the cheap
path, confirming that it reserves spending for statistically ambiguous cases.
The stripped ablations bound the two ends: statistics with no recommendation
mapping (stats\_only) matches the verify floor at 0.75, while a single-shot
judge with no debate (direct\_llm) collapses to 0.50, near chance. Neither
statistics alone nor a lone judge suffices; the structure that routes between
them is what does the work.

\subsection{Stratified behaviour}
Decomposing the aggregate by diversity stratum shows where the difficulty lies.
The low-diversity stratum is easy: every low-diversity corpus fails the gold's
diversity threshold and should be rejected, so there are no positive cases, the
gate rejects all 30 correctly (1.00 accuracy), and $F_1$ is undefined by
construction rather than by failure. The real test is the high-diversity
stratum, where buy and reject cases coexist; there the gate reaches 0.80
accuracy and 0.83 $F_1$. We report the split so the headline 0.90 is read
correctly---as an average over an easy half and a hard half, not a uniform
number.
\begin{table}[t]
  \centering
  \caption{Stratified gate accuracy. The \textsf{dlow} stratum contains no
  positive (buy) cases, so its $F_1$ is undefined-by-construction rather than
  a failure.}
  \label{tab:stratified}
  \begin{tabular}{@{}l S S c@{}}
    \toprule
    Stratum & {Acc.} & {$F_1$} & $n$ \\
    \midrule
    dlow  & 1.000 & 0.000 & 30 \\
    dhigh & 0.800 & 0.833 & 30 \\
    \bottomrule
  \end{tabular}
\end{table}

\subsection{Calibration}
\begin{figure}[t]
  \centering
  \includegraphics[width=\linewidth]{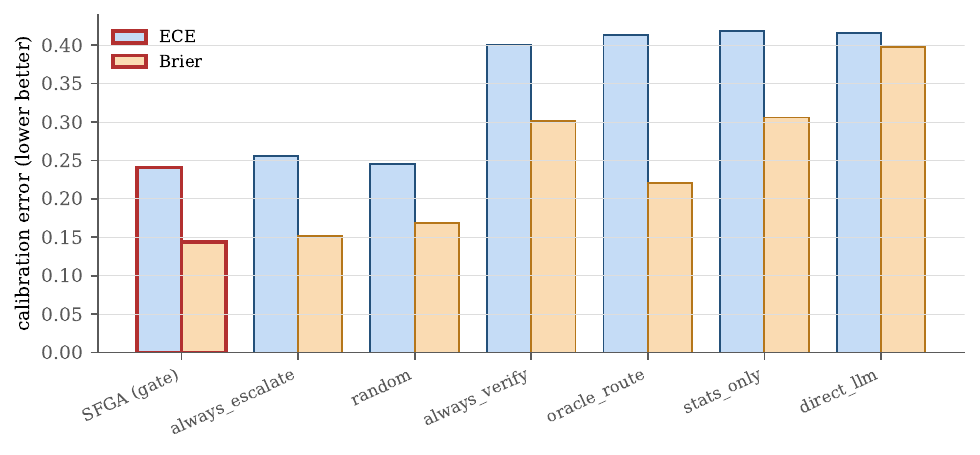}
  \caption{Calibration error (ECE and Brier; lower is better) per strategy. The
  gate (highlighted) is the best calibrated, ahead of the always-verify floor
  and even of oracle-route.}
  \label{fig:calibration}
\end{figure}
Correct routing is not enough for procurement, where a buyer acts on a
probability rather than a bare label. The gate's self-reported buy-probability
is the best calibrated of all strategies (ECE 0.241, Brier 0.144;
\Cref{fig:calibration}, with exact values in \Cref{tab:calibration}). It is
better calibrated than oracle-route (ECE 0.413), which routes perfectly using
the gold yet reports overconfident probabilities, and than the always-verify
floor (ECE 0.400). Blending tight statistics on easy cases with an adjudicated
probability on hard ones yields confidence that tracks correctness better than
either path does alone.

\begin{table}[t]
  \centering
  \caption{Calibration of self-reported buy-probability (exact values for
  \Cref{fig:calibration}).}
  \label{tab:calibration}
  \sisetup{table-format=1.3}
  \begin{tabular}{@{}l S S c@{}}
    \toprule
    Strategy & {ECE} & {Brier} & $n$ \\
    \midrule
    \sys{} (gate)         & \bfseries 0.241 & \bfseries 0.144 & 60 \\
    always\_escalate      & 0.255 & 0.152 & 60 \\
    random\_cost\_matched & 0.245 & 0.169 & 60 \\
    always\_verify        & 0.400 & 0.302 & 60 \\
    oracle\_route         & 0.413 & 0.221 & 60 \\
    stats\_only           & 0.418 & 0.306 & 36 \\
    direct\_llm           & 0.415 & 0.398 & 36 \\
    \bottomrule
  \end{tabular}
\end{table}

\subsection{Debate diagnostics (honest negatives)}
\label{subsec:debate}
\begin{figure}[t]
  \centering
  \includegraphics[width=\linewidth]{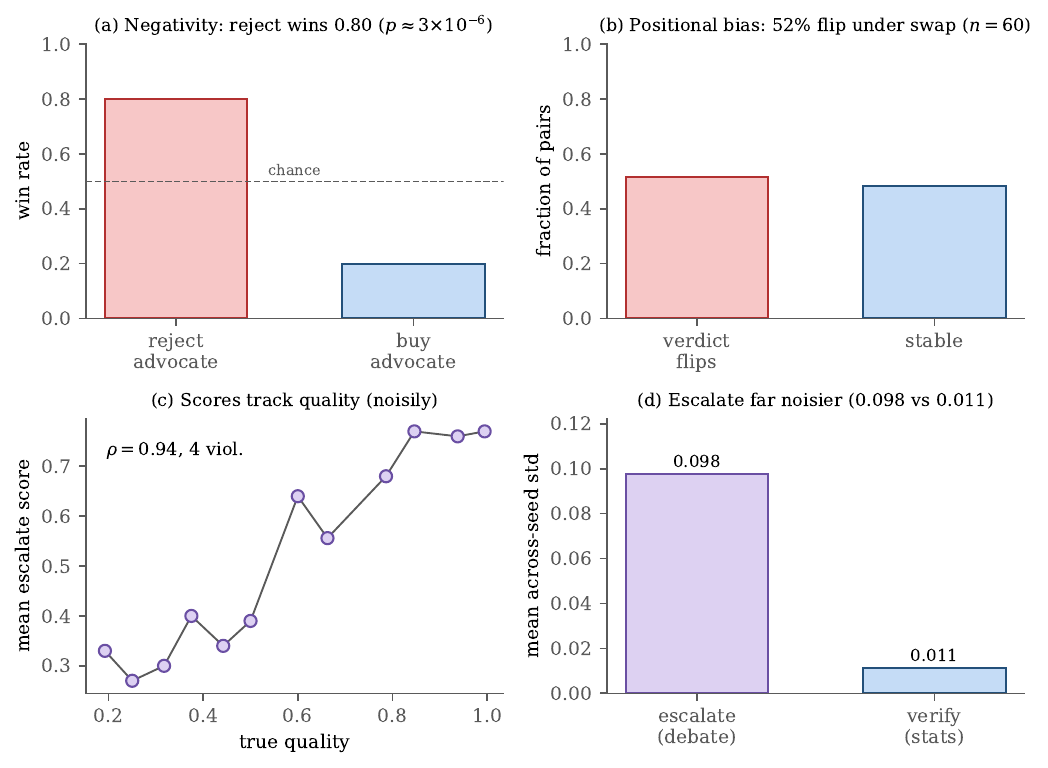}
  \caption{Diagnostics of the escalation path: (a) the reject advocate wins
  most debates; (b) verdicts flip on about half of position-swapped pairs;
  (c) escalate scores still track true quality, but noisily; (d) they are far
  less stable across seeds than verify scores.}
  \label{fig:debate}
\end{figure}
Auditing the escalation path is where we report against our own method
(\Cref{fig:debate}). The
reject advocate wins 0.80 of debates ($p\approx3\times10^{-6}$), a clear
negativity skew; and swapping the advocates' presentation order flips the
verdict on 52\% of the 60 paired cases, a strong positional bias---precisely
the failure modes documented for LLM judges~\citep{wang2024notfair}. These are
not fatal to the design. The escalate scores still track true quality
(Spearman 0.94, with four adjacent-rank violations), and they are far noisier
across seeds than the verify scores (mean standard deviation 0.098 vs.\ 0.011),
which is the empirical signature of a path that should be used sparingly. This
is the core argument for a statistics-first architecture: the adjudication path
is useful but unreliable, so the gate is built to keep it off the critical
route for every case the cheap statistics can already settle
(\Cref{sec:design}).

\section{Discussion}
\label{sec:discussion}

\paragraph{Implications.}
The central finding is that spending should be routed, not scaled. On this
bench a statistics-first gate captures nearly all of the oracle's routing value
(0.90 vs.\ 0.98 accuracy) while consulting the expensive adjudication path on
only the cases cheap statistics cannot settle---and it does so at lower cost
than escalating every case, which is not merely wasteful but actively worse
because it exposes clear-cut cases to a biased judge. The lesson generalizes
beyond this system: when an accurate-but-expensive evaluator has known
pathologies, a cheap decisive front end that gates access to it can be better,
on both cost and accuracy, than using the expensive evaluator everywhere.

\paragraph{Practitioner guidelines.}
The gate exposes a small number of interpretable knobs that trade budget for
caution. Widening the maximum interval width $w_{\max}$ or lowering the
decisive fraction $f$ sends more cases to the cheap path, cutting cost at the
risk of committing on thin evidence; tightening them, or widening the
borderline band $m$, escalates more and buys accuracy with money. Because the
adjudication path carries a measurable negativity skew, a buyer who cares more
about missing good corpora than about admitting weak ones should treat a
reject verdict from debate as weaker evidence than a buy verdict, and the
calibrated probability---not the label---should drive the acquisition
decision.

\paragraph{Limitations.}
Three limits bound our claims. First and most important, the evaluation is
closed-loop: the gold label is a deterministic function of the injected knobs,
and the blind measurements recover those same knobs, so our accuracy figures
demonstrate internal measurement fidelity and routing calibration, not that a
gate-approved corpus actually trains a better model. We have been careful
throughout to frame the results this way. Second, the study uses a single base
corpus (Dolly-15k) and a single judge model family, so transfer across corpora
and model families is untested. Third, the low-diversity stratum contains no
positive cases, so part of the aggregate accuracy comes from an easy
all-reject regime, as the stratified analysis makes explicit.

\section{Conclusion and Future Work}
\label{sec:conclusion}
We presented \sys{}, a statistics-first gate that treats SFT data procurement
as cost-aware routing: cheap interval-valued measurements of diversity,
utility, and redundancy decide most cases directly, and only ambiguous,
borderline, or conflicting ones are escalated to an adjudicative debate. On a
controlled benchmark the gate approaches oracle accuracy at a fraction of the
always-escalate detour, is the best-calibrated strategy tested, and---rather
than hide the debate path's behaviour---quantifies its negativity and
positional biases, turning the case for keeping such a path off the critical
route into a measured argument.

The clearest next step follows directly from the main limitation: replace the
knob-derived gold with a real short SFT run and a held-out evaluation, so that
correctness is defined by whether a gate-approved corpus actually improves a
model, breaking the closed loop. Beyond that, adding a second base corpus and a
second judge family would test whether the routing behaviour and the measured
biases transfer, and richer escalation protocols---more debate rounds, or an
aggregation over swapped orderings that cancels positional bias---may narrow
the remaining gap to the oracle.

\bibliographystyle{plainnat}
\bibliography{references}

\appendix
\section{Hyperparameters and gate configuration}
\begin{table}[H]
  \centering
  \caption{Gate and debate configuration.}
  \label{tab:hparams}
  \begin{tabular}{@{}ll@{}}
    \toprule
    Parameter & Value \\
    \midrule
    min sample size $n_{\min}$        & 20 \\
    max CI width $w_{\max}$           & 0.25 \\
    decisive fraction $f$             & 0.67 \\
    borderline margin $m$             & 0.05 \\
    large-$n$ threshold               & 50 \\
    CI $z$                            & 1.96 \\
    diversity threshold $\thr_{\Div}$ & 0.15 \\
    utility threshold $\thr_{\Util}$  & 0.55 \\
    redundancy threshold $\thr_{\Red}$& 0.25 \\
    authenticity threshold            & 0.55 \\
    seeds                             & 5 \\
    grid                              & $2\times3\times2$ (12 datasets) \\
    \bottomrule
  \end{tabular}
\end{table}

\end{document}